\documentclass[10pt]{article}

\usepackage[utf8]{inputenc} 
\usepackage[T1]{fontenc}    
\usepackage{hyperref}       
\usepackage{url}            
\usepackage{booktabs}       
\usepackage{amsfonts}       
\usepackage{nicefrac}       
\usepackage{microtype}      
\usepackage{xcolor}         

\usepackage{lipsum}
\usepackage{todonotes}
\usepackage{amsmath}
\usepackage{subcaption}
\usepackage{natbib}
\usepackage[
    letterpaper,
    textheight=9in,
    textwidth=5.5in,
    top=1in,
    headheight=12pt,
    headsep=25pt,
    footskip=30pt
]{geometry}
\usepackage{times}
\usepackage[parfill]{parskip}

\newcommand\extrafootertext[1]{%
    \bgroup
    \renewcommand\thefootnote{\fnsymbol{footnote}}%
    \renewcommand\thempfootnote{\fnsymbol{mpfootnote}}%
    \footnotetext[0]{#1}%
    \egroup
}

\title{Promoting User Data Autonomy During the Dissolution of a Monopolistic Firm}

\author{
    \fontsize{10}{12}\selectfont
    \begin{tabular}[t]{@{}c@{\hskip 1in}c@{}}
        \textbf{Rushabh Solanki} & \textbf{Elliot Creager} \\
        University of Waterloo, Vector Institute & University of Waterloo, Vector Institute \\
        \texttt{r7solank@uwtaerloo.ca} & \texttt{creager@uwaterloo.ca}
    \end{tabular}
}

\date{}

\begin{document}

\maketitle

\extrafootertext{This paper appeared at the 2nd Workshop on Regulatable ML at NeurIPS 2024.}

\begin{abstract}
The deployment of AI in consumer products is currently focused on the use of so-called foundation models, large neural networks pre-trained on massive corpora of digital records. This emphasis on scaling up datasets and pre-training computation raises the risk of further consolidating the industry, and enabling monopolistic (or oligopolistic) behavior. Judges and regulators seeking to improve market competition may employ various remedies. This paper explores \emph{dissolution}---the breaking up of a monopolistic entity into smaller firms---as one such remedy, focusing in particular on the technical challenges and opportunities involved in the breaking up of large models and datasets. We show how the framework of Conscious Data Contribution can enable user autonomy during under dissolution. Through a simulation study, we explore how fine-tuning and the phenomenon of “catastrophic forgetting” could actually prove beneficial as a type of machine unlearning that allows users to specify which data they want used for what purposes.

\end{abstract}

\section{Introduction}

In August 2024, the D.C. District Court delivered a landmark ruling, finding that Google had engaged in anti-competitive practices in its search engine business, in violation of the Sherman Act~\citep{goldstein_will_2024}.
This ruling signals an increasing willingness by the US Department of Justice to challenge the concentration of power in the technology sector.
It also coincides with the waxing influence of AI-powered software on consumer markets, where the increasing reliance on so-called foundation models could further exacerbate competition issues~\citep{bommasani_opportunities_2022,brynjolfsson_turing_2022}.

The Google anti-trust case is far from resolved: pending an ongoing appeal, it will conclude with judges suggesting practical ``remedies,'' such as fines or restrictions on business practices, to encourage market competition going forward.
This paper explores the hypothetical use of a particularly stern remedy to address monopolistic behavior by an AI firm: the \emph{dissolution} of a offending firm into smaller successor companies. 
While dissolution was employed against 20th century monopolists such as Standard Oil and AT\&T, it has not been used 
since the advent of the digital economy.

The potential for dissolution of AI firms is muddled by mirky technical questions such as how best to ``break up'' the large datasets and models of a monopolist firm.
We note that the dissolution of an AI firm (or any serious corporate restructuring imposed by a regulator) presents the chance to rethink standard data curation practices that emphasize scale over consent~\citep{hanna_against_2020,andreotta_ai_2022}.
We advocate for the importance of data autonomy in the dissolution process, drawing on the framework of Conscious Data Contribution~\citep{vincent_can_2021} to ensure that users can have say in how their data, previously owned by the monopolist, is used by the successor firms going forward.
Using a simple and stylized simulation study, we explore the possibility of realizing dissolution that respects user data autonomy using standard fine-tuning practices.

\section{Background}

\paragraph{Data Autonomy}
Our explorations into regulating AI during dissolution emphasize user data autonomy, and are inspired by Data Leverage~\citep{vincent_data_2021}. 
This framework was originally introduced to broadly describe how users can exert indirect influence over AI systems by modifying their data.
We focus in particular on Conscious Data Contribution (CDC) \citet{vincent_can_2021}---a particular data lever whereby users choose to share their data with smaller firms---as a way to ensure users' data consent is respected when dissolution is 
carried out by regulators.

\paragraph{Model Adaptation}
Prior work on CDC focused on how users' data contributions have a disproportionately positive impact for smaller firms who train their AI models from scratch~\citep{vincent_can_2021}.
Because we focus instead on applying the CDC principle to \emph{pre-trained} foundation models owned by the monopolistic firm, our work also relates to other approaches seeking to adapt large models.
Model merging seeks to combine multiple pre-trained models directly in their parameter space, thus avoiding the need to retrain from scratch when the source dataset changes~\citep{ainsworth_git_2023,matena_merging_2022}.
By analogy, this might represent the type of technique suitable for the merger of two firms, rather than the dissolution of a single firm.

Removing the influence of specific data points from a model is the focus of Machine Unlearning~\citep{bourtoule_machine_2021}, motivated in part by the ``right to be forgotten'' principle within the EU GDPR~\citep{wachter_counterfactual_2018}. Unlearning is difficult to implement due to an inherent tension between forgetting the target data point and preserving the model's fit to the remaining distribution~\citep{kurmanji_towards_2023}; accordingly, contemporary unlearning schemes often still rely on retraining from scratch. On the other hand, there is evidence that fine-tuning a model on a new task  induces \emph{catastrophic forgetting}, a type of natural unlearning resulting from the collapse of the model's representation to retain only features relevant to the new task~\citep{french_catastrophic_1999}.
Below we discuss how this phenomenon makes fine-tuning a simple and appealing approach for implementing CDC under dissolution.
Moreover, existing machine unlearning methods are primarily aimed at removing specified objects, patterns or stylistic elements, rather than addressing varied diverse data as in our case -- making fine-tuning an even more attractive option.

\paragraph{AI and Market Competition}
The heavy emphasis on scaling up data and compute power to build modern AI products raises concerns over the ability for smaller firms to compete~\citep{brynjolfsson_turing_2022}.
\citet{luitse_great_2021} analyze the political economy of the transformer architecture, including the tendency of transformer-based LLMs to consolidate power within the tech industry.
\citet{burkhardt_foundation_2024} discuss how the focus of prompting as an interface for foundation models further supports the recent \emph{platformization} of the software industry~\citep{helmond_platformization_2015}, possibly strengthening the market hold of larger firms.
Recently, simulation studies have suggested that automated AI decision making in the domain of pricing algorithms can unexpectedly lead to collusive behavior, both for models based on Q-learning~\citep{calvano_artificial_2020} and LLMs~\citep{fish_algorithmic_2024}.

\section{Conscious Data Contribution Under Firm Dissolution}

We consider a scenario in which a monopolistic firm $F$ is dissolved into $k$ successor companies $\{S_1, \ldots, S_k\}$. Before dissolution, $F$ owns a dataset $U \in \mathbb{R}^{N \times d}$ of $d$-dimensional data records from each of its $N$ users. We denote $u_i \subset U$ as the data about the $i$-th user within the dataset, and further assume that this user data is a vector $u_i \in \mathbb{R}^d$ built by concatenating $J$ task-specific user data vectors:
\[
u_i = \begin{pmatrix}
u_i^{(1)} \\
u_i^{(2)} \\
\vdots \\
u_i^{(J)}
\end{pmatrix},
\]
where $u_i^{(j)} \in \mathbb{R}^{d_j}$ and $\sum_j d_j = d$.
For example, $u_i^{(1)}$ might correspond to location data collected about user $u_i$, while  $u_i^{(2)}$ might correspond to that user's music recommendation data. 

As the monopolistic firm $F$ is dissolved into successor firms  $\{S_1, \ldots, S_k\}$, the key regulatory question is what will happen to the dataset $U$, and any models derived from it.
To allow successor firms to build expressive models in a sample efficient manner, while preserving data autonomy, 
we posit an avenue for Conscious Data Contribution~\citep{vincent_can_2021} based on fine-tuning. 
We introduce a binary matrix for each user $C(u_i) \in \{0, 1\}^{K \times J}$, whose entries correspond to that user consciously contributing task-specific data to particular successor firms.
In other words, we have
\[
[C(u_i)]_{j,k} = 
\begin{cases}
1 & \text{if user \emph{i} contributes their task-\emph{j} data to successor \emph{k},} \\
0 & \text{else.}\\
\end{cases}
\]

For example, if $S_k$ represents the successor company taking on the music streaming business from $F$, then user $U_i$ may wish to set $[C(u_i)]_{1,k}=1$ in order to contribute their music listening history to that successor company, while setting  $[C(u_i)]_{2,k}=0$ to preclude the use of location data by $S_k$.\footnote{Apple's ``ask app not to track'' iOS feature~\citep{chen_be_2021} can be seen as a realization of this type of CDC where $\{S_1, \ldots S_K\}$ represent (possibly competing) apps rather than successor companies.}  
Thus the $i$-th user consciously contributes the following data vector to the $k$-th successor firm: 
\[
u^{\text{CDC}-S_k}_i = \begin{pmatrix}
u_i^{(1)} \cdot [C(u_i)]_{1,k} \\
u_i^{(2)} \cdot [C(u_i)]_{2,k} \\
\vdots \\
u_i^{(J)} \cdot [C(u_i)]_{2,J} \\
\end{pmatrix}.
\]

Accordingly, $S_k$ receives $U^{\text{CDC}-S_k} = [u^{\text{CDC}-S_k}_1, \ldots, u^{\text{CDC}-S_k}_N]^T$, a dataset of consciously contributed data for the purposes of training their models, with non-consented entries zeroed out.

Through $C(u_i)$, each user has the opportunity to consciously contribute portions of their data (organized by task-appropriateness) to each successor company. 
In order for this scheme to promote user data autonomy within a competitive market, regulators would need to establish a framework for ensuring that users' wishes are respected by the firms. 
In particular, this introduces the following two regulatory challenges:
\begin{enumerate}
    \item Firms should only gain access to data permitted under $C(u_i)$;
    \item When $[C(u_i)]_{j,k} = 0$, reasonable steps should be taken to ensure that information about user $i$ is not used by firm $k$ to solve task $j$. 
\end{enumerate}

Satisfying these requirements is complicated for firms that rely on foundation models: the successor companies may not have sufficient data within  $U^{\text{CDC}-S_k}$ to train a performant model from scratch (impeding (1)), and the original foundation model of $F$ may not be directly interpretable or explainable (impeding (2)).
To take a first step towards realizing a CDC approach for foundation models, we next examine a stylized setting focused on the dynamics of fine-tuning smaller neural networks, where we find the common phenomenon of catastrophic forgetting promotes user data autonomy while allowing the firms a sample-efficient way to comply with 
these regulatory challenges.

\section{Simulation Studies}

We conduct proof-of-concept simulations across two data modalities. 
Upon dissolution, we suppose a successor firm $S_k$ can preserve only a portion of the original data obtained through Conscious Data Contribution, that is $U^{\text{CDC}-S_k}$, but may fine tune some model previously trained on $U$.
For simplicity we assume $k = 1$, then denote $U^{\text{CDC}}$ as the available data to the single successor firm and $\neg{U}^{\text{CDC}} := U - U^{\text{CDC}}$ as the data the successor needs to unlearn.

\subsection{Datasets}

\paragraph{Image generation} We begin with a simple setting where the firm's data $U$ was used to train a class-conditional diffusion model.
We carry out the same experimental procedure first on MNIST and then on CIFAR-10.
To keep it simple, we suppose the sole successor $S$ receives the data $U^{\text{CDC}}$ comprising three of the ten classes. 
While the original diffusion model can generate images that match its conditioned class, if user autonomy is respected then the model fine-tuned with $U^{\text{CDC}}$ will produce samples according to the new distribution, even if conditioned on classes outside of $U^{\text{CDC}}$.

\paragraph{Text classification}
We train a DistillBERT model to predict which sets of skills are described in a dataset of $30,000$ resumes from a major gig labor platform~\citep{jiechieu2021skills}.
Importantly, this is multi-label classification as skills may not be unique.
We consider a single employer $F$ (hiring from all skills) that is dissolved into one successor $S$ hiring from a single skill set, so $U^{\text{CDC}}$ as all resumes with ``administrative'' skills and $\neg U^{\text{CDC}}$ as all resumes with ``developer'' skills.
As before, $S$ may optionally fine-tune on $U^{\text{CDC}}$, starting from $F$'s model previously trained on $U$.

\subsection{Results}
The successor $S$ is satisfied if they have a sample efficient way to produce a model that generalizes well to unseen data from $\mathbb{P}(U^{\text{CDC}})$, while the users and regulator are satisfied if this model generalizes poorly to $\mathbb{P}(\neg U^{\text{CDC}})$.
Thus we measure the ``retain rate'' on held out data from $U^{\text{CDC}}$ and ``forget rate'' on data from $\neg U^{\text{CDC}}$, where both metrics are ``the higher the better''.
For the image generation task, given a snapshot of a trained diffusion model, 10K images were sampled comprising 1000 samples per class. The forget rate was measured for 7K samples corresponding to $\neg{U}^{\text{CDC}}$ classes using 1 minus accuracy, and we used accuracy for the remaining 3K samples.
For the resume dataset, we instead use $F1$ score of the DistillBERT classifier evaluated on successor data for retain rate, and $1$ minus $F1$ score on non-successor data for forget rate.
See Appendix~\ref{apdx:exp_details} for details.

\begin{table}[!t]
  \caption{Main Results for Diffusion Experiments}
  \vspace{-0.1cm}
  \label{tab:main_results_diff}
  \centering
\resizebox{0.95\textwidth}{!}{
  \begin{tabular}{ccccccc}
    \toprule
     & \multicolumn{2}{c}{MNIST @ 100 epochs} & \multicolumn{3}{c}{CIFAR10 @ 100 epochs} \\
     \cmidrule(lr){2-3}\cmidrule(lr){4-6}
     & Forget Rate on $\neg U^{\text{CDC}}$ ($\uparrow$) & Retain Rate on $U^{\text{CDC}}$ ($\uparrow$) & Forget Rate on $\neg U^{\text{CDC}}$ ($\uparrow$) & Retain Rate on $U^{\text{CDC}}$ ($\uparrow$) & FID scores ($\downarrow$) \\
    \midrule
    Original  & 0.016 & 0.97 & 0.381 & 0.544 & 71.013  \\
    \midrule
    Retraining & 0.962 & 0.955 & 0.918 & 0.266 & 132.00 \\
    Fine-tuning & 0.955 & 0.914 & 0.844 & 0.536 & 57.550 \\
    \bottomrule
  \end{tabular}
}
\end{table}

\begin{table}[!t]
  \caption{Main Results for Resume Dataset (@ 25 epochs)}
  \vspace{-0.1cm}
  \label{tab:main_results_res}
  \centering
  \scalebox{0.75}{
  \begin{tabular}{ccccccc}
    \toprule
     & Forget Rate (F1) on $\neg U^{\text{CDC}}$ ($\uparrow$) & Retain Rate (F1) on $U^{\text{CDC}}$ ($\uparrow$) \\
    \midrule
    Original  & - & 0.898  \\
    \midrule
    Retraining (pre-trained) & 0.937 & 0.901  \\
    Retraining (scratch) & 0.946 & 0.805  \\
    Fine-tuning & 0.917 & 0.964  \\
    \bottomrule
  \end{tabular}
  }
\end{table}

Our proposed approach is to fine-tune on $U^{\text{CDC}}$ starting from a model pre-trained on $U$, which we compare with two baselines: training on $U^{\text{CDC}}$ from scratch, and fine-tuning from a domain-agnostic pre-training initialization (e.g. huggingface weights). 
While retraining from scratch works well for small datasets like MNIST, it suffers from poor utility on larger datasets, where we find that fine-tuning more favorably balances successor utility with user data autonomy (Table~\ref{tab:main_results_diff} and Table~\ref{tab:main_results_res}).
We also attempted gradient ascent-based unlearning, but it became unstable after a few epochs when applied to unlearn large data.
These experiments are performed under a simple setting, whereas the process of real-world dissolution will introduce multiple complexities in the distribution of $U^{\text{CDC}}$. While limited in scope, these results show that fine-tuning provides a natural unlearning effect for users' data that was not consciously contributed to the successor, suggesting that promoting users' data autonomy within a complex regulatory schema may be feasible through simple means.

\section{Conclusion}

Dissolving a large firm would be a substantial undertaking, which would almost certainly involve political and legal friction.
For example, in the 1990s monopoly case brought against Microsoft, a dissolution remedy initially explored by the judiciary was ultimately watered down, and the actual remedies realized by the courts were less disruptive to Microsoft's business~\citep{goldstein_will_2024}.
Nevertheless, we feel that the emerging field of AI regulation should explore frameworks that account for all possible remedies available to counteract monopolistic behavior. 
In this paper we have taken one step towards exploring a regulatory framework for dissolution through the lens of Conscious Data Contribution~\citep{vincent_can_2021}, with the aims of ensuring that when a monopolistic firm is broken up, its successor firms only use data and models with explicit user consent.
Through a simple proof-of-concept simulation study we demonstrated that the phenomenon of catastrophic forgetting~\citep{french_catastrophic_1999} during fine-tuning promotes user autonomy in this case.

\section*{Acknowledgments}

We are grateful to Nicholas Vincent and the anonymous reviewers for their constructive feedback.
Resources used in preparing this research were provided, in part, by the Province of Ontario, the Government of Canada through CIFAR, and companies sponsoring the Vector Institute~\url{www.vectorinstitute.ai/partnerships/.}

\medskip

{
    \small
    \bibliographystyle{plainnat}
    \bibliography{refs}
}


\newpage
\appendix

\section{Experimental Details}
\label{apdx:exp_details}

\paragraph {Image generation} Our conditional diffusion model was implemented using the Hugging Face's Diffusers library. We utilized the standard configuration, which uses the DDPM scheduler with 1000 time steps and a cosine variance schedule. In all the experiments, we have used the batch size of 128. For MNIST data, we used the UNet architecture, with approximately 1.71M trainable parameters, which was trained for 46.9K parameter updates. Whereas the model trained on the CIFAR10 dataset has around 6.34M parameters and was trained for 195.5K parameter updates. We chose the ADAM optimizer with a learning rate 0.001 and used Mean Squared Error (MSE) as the loss function for both datasets.

The MNIST classifier used to compute the forget rate was trained from scratch with PyTorch's ResNet18 implementation for roughly 164K parameter updates.
Meanwhile, the CIFAR10 classifier was fine-tuned from pre-trained ImageNet weights using a ResNet50 model, with around 39K parameter updates. Test set accuracies for both datasets are provided in Table \ref{tab:test_acc}.

\begin{table}[!h]
  \caption{The test performance of both the classifiers}
  \label{tab:test_acc}
  \centering
  \begin{tabular}{ccc}
    \toprule
    Dataset & Test Accuracy \\
    \midrule
    MNIST & 0.997  \\
    CIFAR10 & 0.969 \\
    \bottomrule
  \end{tabular}
\end{table}

\paragraph{Text classification} We used pre-trained \texttt{distilbert-base-uncased} from HuggingFace transformers library corresponding to the DistilBERT transformer~\citep{sanh2020distilbert}. After preprocessing, the dataset contains a total of 29020 samples, split into 25000 for training and 4020 for testing. 
Following dissolution, $U^{\text{CDC}}$ holds 14009 samples, divided into a training set of 11907 and a test set of 2102, while the remaining samples are treated as $\neg{U}^{\text{CDC}}$.

\begin{figure}[!b]
  \centering
  \hspace{-1.5cm}
  \begin{subfigure}[b]{0.45\textwidth}
    \centering
    \includegraphics[scale=0.475]{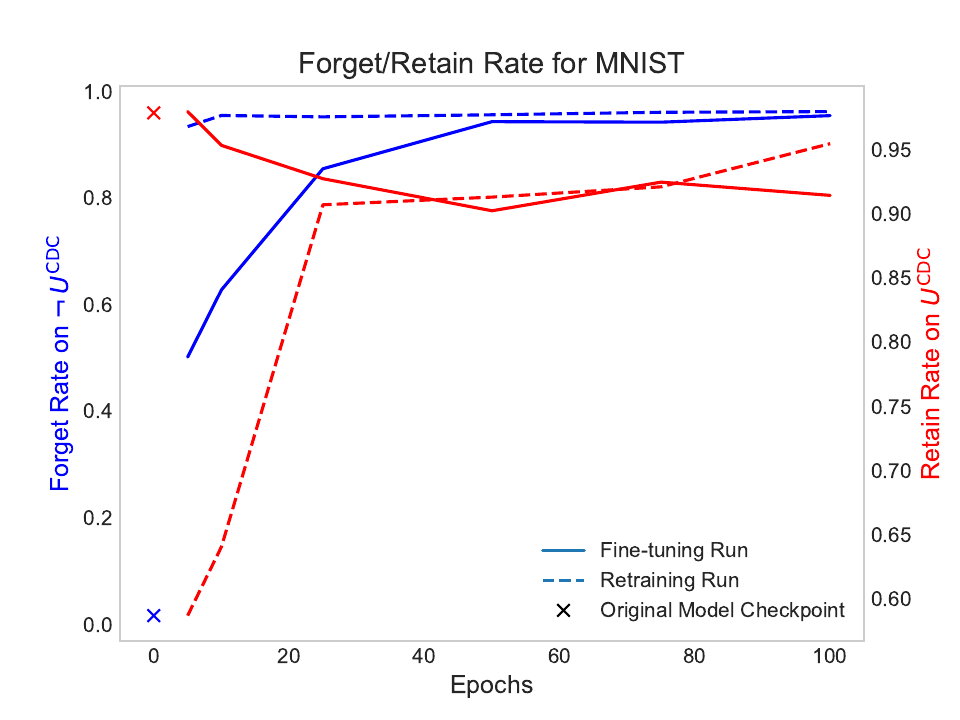}
    \vspace{-0.2cm}
  \end{subfigure}
  \hspace{1.25cm}
  \begin{subfigure}[b]{0.45\textwidth}
    \centering
    \includegraphics[scale=0.475]{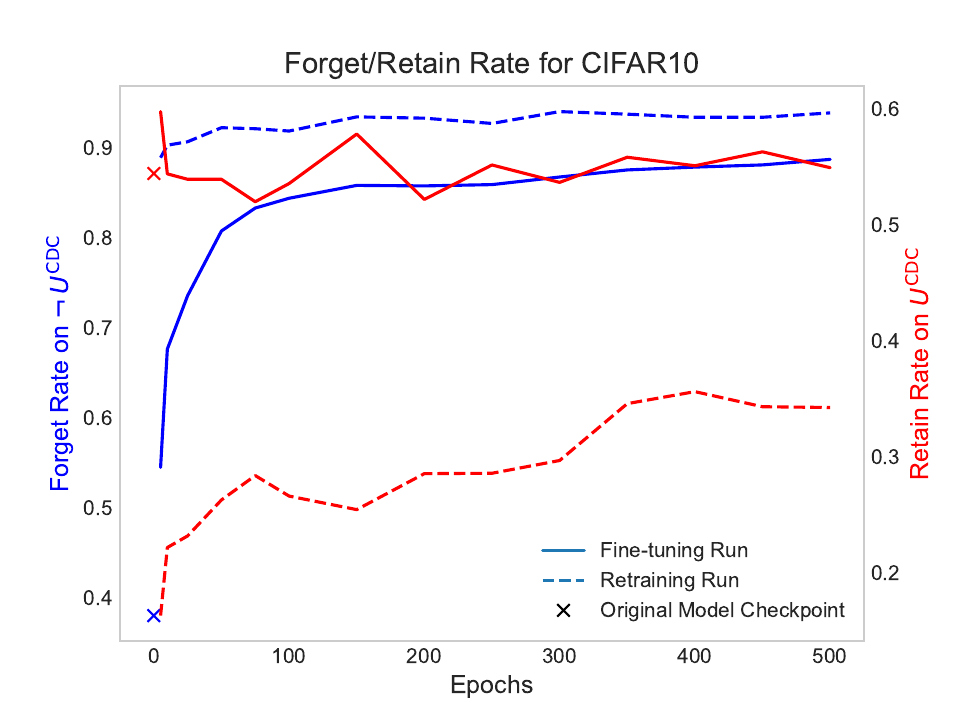}
    \vspace{-0.2cm}
  \end{subfigure}
  \\
  \vspace{-0.25cm}
  \begin{subfigure}[b]{1\textwidth}
    \centering
    \includegraphics[scale=0.475]{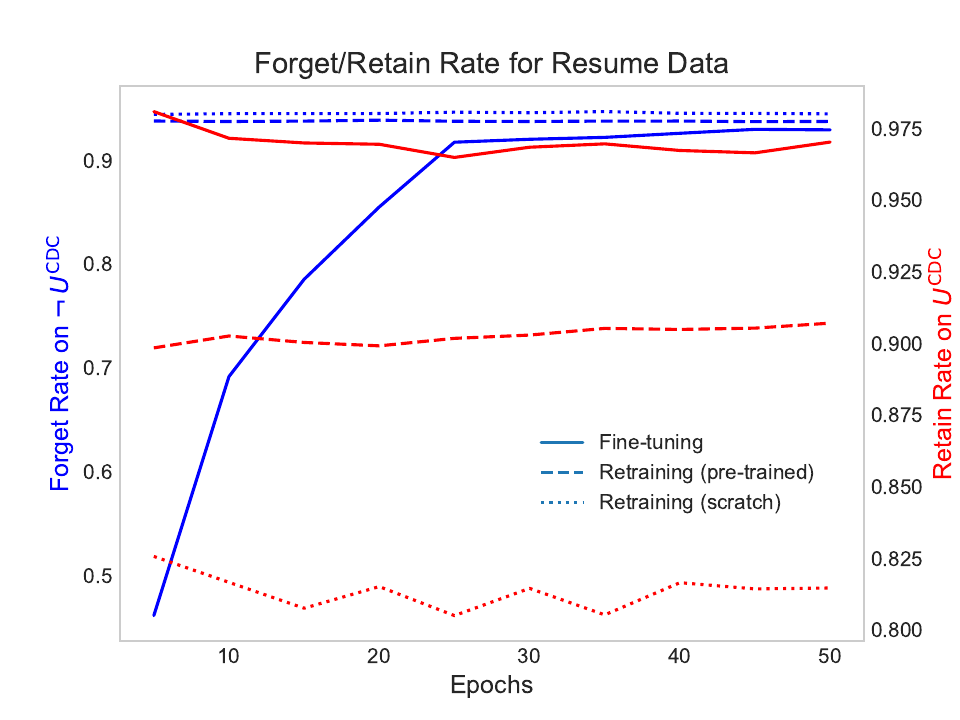}
    \vspace{-0.2cm}
  \end{subfigure}
  \caption{The plots illustrates the changes in forget and retain rates during the process of fine-tuning and retraining with consciously contributed data $U^{\text{CDC}}$ to a successor $S$ of a firm upon \emph{dissolution}. Each plot displays the results for different datasets, with the top two for image generation and bottom one for text classification.}
  \label{fig:forget_rate}
\end{figure}

Figure \ref{fig:forget_rate} shows the increase in the forget rate as the fine-tuning progresses with almost 0\% at the start of fine-tuning representing the base model. For MNIST data, at only the $25^{th}$ epoch of fine-tuning, we get the forget rate of 91.6\%. Attaining a comparable forget rate for the CIFAR10 involved a considerable number of epochs and plateaus of around 100 epochs. The forget rate for the resume data converges at around 25 epochs and overall fine-tuning generalizes well when comparing the retain rates.

\end{document}